\documentclass[11pt,a4paper]{article}
\usepackage[hyperref]{naaclhlt2018}
\usepackage{times}
\usepackage{latexsym}
\usepackage{graphicx}
\usepackage[normalem]{ulem}
\useunder{\uline}{\ul}{}
\usepackage{amsmath}
\makeatletter
\let\@fnsymbol\@arabic
\makeatother
\usepackage{url}

\aclfinalcopy % Uncomment this line for the final submission
\title{Building a Conversational Agent Overnight with Dialogue Self-Play}

\author{Pararth Shah\thanks{\hspace{0.1cm} Correspondence to pararth@google.com}, Dilek Hakkani-T{\"u}r, Gokhan T{\"u}r, Abhinav Rastogi,
\\ {\bf Ankur Bapna, Neha Nayak, Larry Heck}
\\ Google AI \\ Mountain View, CA, USA}

\date{}

\begin{document}
\maketitle
\begin{abstract}
We propose Machines Talking To Machines (M2M), a framework combining automation and crowdsourcing to rapidly bootstrap end-to-end dialogue agents for goal-oriented dialogues in arbitrary domains. M2M scales to new tasks with just a task schema and an API client from the dialogue system developer, but it is also customizable to cater to task-specific interactions. Compared to the Wizard-of-Oz approach for data collection, M2M achieves greater diversity and coverage of salient dialogue flows while maintaining the naturalness of individual utterances. In the first phase, a simulated user bot and a domain-agnostic system bot converse to exhaustively generate dialogue ``outlines'', i.e. sequences of template utterances and their semantic parses. In the second phase, crowd workers provide contextual rewrites of the dialogues to make the utterances more natural while preserving their meaning. The entire process can finish within a few hours. We propose a new corpus of 3,000 dialogues spanning 2 domains collected with M2M, and present comparisons with popular dialogue datasets on the quality and diversity of the surface forms and dialogue flows.
\end{abstract}

\section{Introduction}
\begin{figure}[ht!]
  \includegraphics[width=0.99\linewidth]{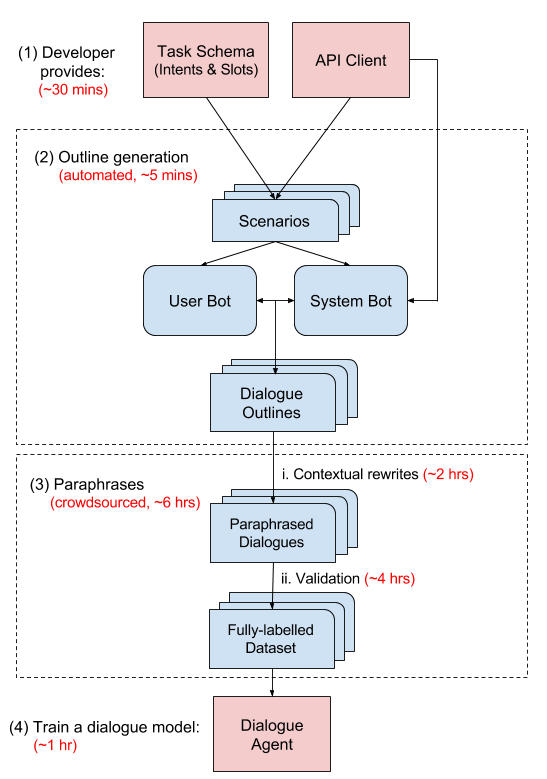}
  \caption{Our proposed M2M framework: (1) the dialogue developer provides a task schema and an API client, (2) automated bots generate dialogue outlines, (3) crowd workers rewrite the utterances and validate slot spans, (4) a dialogue model is trained with supervised learning on the dataset. The whole process can complete in under 8 hours.
  }
  \label{fig:process}
\end{figure}
Goal-oriented dialogue agents trained using supervised learning methods work best when trained on dialogues of the same task. However, when developing a dialogue agent to assist a user for completing a new task, for example scheduling a doctor’s appointment via an online portal, a dataset of human-agent dialogues for that task may not be available since no dialogue agent exists for interacting with that particular API. One popular approach is to collect and annotate free-form dialogues via crowdsourcing using a Wizard-of-Oz setup (\citet{wen2016network, asri2017frames}). However, this is an expensive and lossy process as the free-form dialogues collected from crowdworkers (i) might not cover all the interactions that the agent is expected to handle, (ii) might contain dialogues unfit for use as training data (for instance if the crowd workers use language that is either too simplistic or too convoluted), and (iii) may have errors in dialogue act annotations, requiring an expensive manual filtering and cleaning step by the dialogue developer.

Another approach, popular among consumer-facing voice assistants, is to enable third-party developers to build dialogue ``experiences'' or ``skills'' focusing on individual tasks (e.g. DialogFlow\footnote{https://dialogflow.com}, Alexa Skills\footnote{https://developer.amazon.com/alexa-skills-kit}, wit.ai\footnote{https://wit.ai}). This provides the dialogue developer with full control over how a particular task is handled, allowing her to incrementally add new features to that experience. However, this approach relies heavily on the developer to engineer every aspect of the conversational interaction and anticipate all ways in which users might interact with the agent for completing that task. It is desirable to expand this approach to make it more data-driven, bringing it closer to the Wizard-of-Oz approach popular in the dialogue research community.

We present Machines Talking To Machines (M2M), a functionality-driven process for training dialogue agents. The primary goal is to reduce the cost and effort required to build dialogue datasets by automating the task-independent steps so that a dialogue developer is required to provide only the task-specific aspects of the dialogues. Another goal is to obtain a higher quality of dialogues in terms of: (i) diversity of language as well as dialogue flows, (ii) coverage of all expected user behaviors, and (iii) correctness of supervision labels. Finally, this framework is aimed towards bootstrapping dialogue agents up to the point where they can be deployed to serve real users with an acceptable task completion rate, after which they should be improved directly from user feedback using reinforcement learning.

Previous work for building semantic parsers (\citet{wang2015building}) and parsers for mapping natural language questions to structured queries (\citet{zhong2017seq2sql}) rely on crowd sourcing to map automatically generated structured representations to single-shot natural language utterances. However, generating multi-turn dialogues in this manner requires co-ordination among multiple participating agents. Inspired by recent AI game-playing literature (\citet{silver2016mastering,silver2017mastering}), we introduce a notion of ``dialogue self-play'' where two or more conversational agents interact by choosing discrete conversational actions to exhaustively generate dialogue histories. In this work, we employ an agenda-based user simulator agent (\citet{schatzmann2007agenda}) and a finite state machine based system agent for the self-play step.

In Section 2 we describe the mechanics of M2M and in Sections 3 and 4 we discuss the user simulation and crowdsourcing aspects of our method. In Section 5 we present datasets collected with this framework that we are releasing with this paper and in Section 6 we evaluate our approach by comparing our datasets with popular dialogue datasets. We conclude with a discussion of related work in Section 7.

\section{M2M}
At a high level (Figure \ref{fig:process}), M2M connects a \textit{developer}, who provides the task-specific information, and a \textit{framework}, which provides the task-independent information, for generating dialogues centered around completing the task. Formally, the framework $F$ maps a task specification $T$ to a set of dialogues $D$:

\begin{align}
F(T) &\rightarrow D = \left \{ d_i, i \in 1 \dots N \right \} \\
d_i &= [(u^i_1, \dots, u^i_{n_i}), (a^i_1, \dots, a^i_{n_i})]
\end{align}

Each dialogue $d_i$ is a sequence of natural language utterances (or \textit{dialogue turns}) $u^i_j$ and their corresponding annotations $a^i_j$. A dialogue turn annotation $a^i_j$ includes the semantic parse of that turn as well as additional information tied to that turn, for example who spoke at that turn and the dialogue state at that point in the dialogue.

\begin{figure*}
  \includegraphics[width=\textwidth,height=5cm]{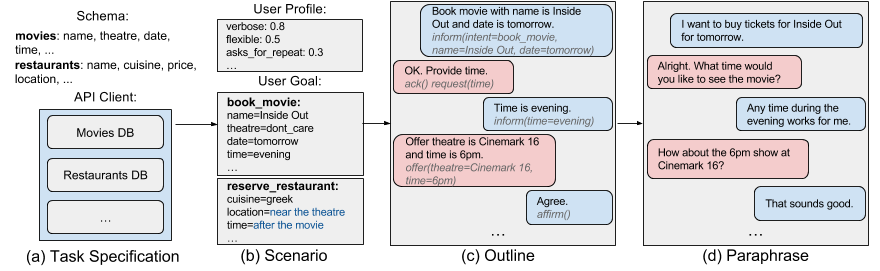}
  \caption{Example of generating an outline and its paraphrase. See text for details.}
  \label{fig:framework}
\end{figure*}

\subsection{Dialogue task specification}
The input to the framework is a task specification obtained from the dialogue developer, which defines the scope of the dialogue interactions for the task in question. Dialogue agents can be employed to complete a wide variety of tasks. In this work we focus on database querying applications, which involve a relational database which contains entities that the user would like to browse and select through a natural language dialogue. This formulation covers a large variety of tasks that are expected of automated dialogue agents, including all tasks that map to filling a form and executing a transaction on some website. The attributes of the entities (i.e. columns of the database) induce a schema $S$ of ``slots''. Each slot could be a constraint that the user cares about when selecting an entity. The developer must provide an API client $C$ which can be queried with a SQL-like syntax to return a list of matching candidate entities for any valid combination of slot values. The task schema and API client together form the task specification, $T = (S, C)$. (Figure \ref{fig:framework}a.)

Dialogues that involve procedural turn-taking (for example reading out a recipe or playing text-based games), or deal with unstructured knowledge (for example question answering over a text document), are among tasks that are not covered by this formulation. These classes of dialogues can be handled by modifying the self-play phase of the framework to generate outlines for these dialogue types.

\subsection{Outline generation via self-play}
With the task specification, the framework must generate a set of dialogues centered around that task. We divide this into two separate steps, $F = F_2 \circ F_1$, where $F_1$ maps the task specification to a set of outlines $O$, and $F_2$ maps each outline to a natural language dialogue:

\begin{align}
F_1(T) &\rightarrow O = \left \{ o_i, i \in 1 \dots N \right \} \\
o_i &= [(t^i_1, \dots, t^i_{n_i}), (a^i_1, \dots, a^i_{n_i})] \\
F_2(\{o_i\}) &\rightarrow \{d_i\}
\end{align}

We define an outline $o_i$ as a sequence of template utterances $t^i_j$ and their corresponding annotations $a^i_j$. Template utterances are simplistic statements with language that is easy to generate with a few rules, as we will describe below. An outline encapsulates the flow of the dialogue while abstracting out the variation in natural language of the surface forms of each dialogue turn. Outlines are easier to generate using self-play between a user bot and a system bot as the bots do not need to generate complex and diverse language that mimics real users and assistants.

To generate an outline, the framework first samples a \textit{scenario} from the task specification. We define a scenario as a user profile and user goals, $s_i = (p_i, g_i)$ (Figure \ref{fig:framework}b). In a goal-oriented dialogue, the user wants to accomplish goals with the assistance of the dialogue agent, for example booking movie tickets or reserving restaurant tables. Each goal is associated with constraints that map to slots of the schema, for example the movie name, genre, number of tickets, and price range. The slots in a user goal can have fixed values (e.g. genre should be ``comedy'' or the user will deny the offer), a list of possible values (e.g. genre should ``comedy'' or ``action''), flexible values (e.g. ``comedy'' is preferred, but the user is open to other options), or open values (e.g. the user is open to seeing movies of any genre). In the multi-domain setting, a goal's slot values can refer to previous goals, for example the user may want to buy a movie ticket and then get dinner after the movie at a restaurant near the theatre chosen in the preceding sub-dialogue. A \textit{scenario generator} samples goals $g_i$ from the task specification by randomly choosing the constraint type and values for every slot in the schema. The values are chosen from a set that includes all available values in the database as well as some non-existent values to create unsatisfiable user goals.

In addition to the user goal, the flow of the dialogue is also dependent on the personality of the user. A user could be verbose in specifying more constraints in a single turn, or could prefer to give each constraint separately. Similarly a user could be more or less amenable to changing their goal if their original constraints are not satisfiable. We define a user profile vector $p_i$ to encapsulate all the task-independent characteristics of the user's behavior. In its simplest version, $p_i$ could be modeled as a vector of probabilities concerning independent aspects of the user's behavior, which could be passed into a programmed user simulator. Alternatively, $p_i$ could be an embedding of a user profile in a latent space, which could condition a learned user simulator model. In our setup, the scenario generator samples $p_i$ from a manually specified distribution of typical user profiles.

With the dialogue scenario $s_i$, the framework performs dialogue self-play between a \textit{user bot} $B_U$ and \textit{system bot} $B_S$ to generate a sequence of turn annotations $a^i_1 \dots a^i_{n_i}$ as follows:

\begin{align}
B_U &= P(a^i_j|a^i_1, \dots, a^i_{j-1}, p_i, g_i) \\
B_S &= P(a^i_{j+1}|a^i_1, \dots, a^i_j, S, C)
\end{align}

Each turn annotation $a^i_j$ consists of a \textit{dialogue frame} that encodes the semantics of the turn as a \textit{dialogue act} and a \textit{slot-value map}, for example ``inform(date=tomorrow, time=evening)'' is a dialogue frame that informs the system of the user's constraints for the date and time slots. Table \ref{tab:dacts} in Appendix \ref{sec:supplemental} has a full list of dialogue acts. $B_U$ maps a (possibly empty) dialogue history $a^i_1 \dots a^i_{j-1}$ and a scenario $p_i, g_i$ to a distribution over turn annotations for the next user turn. Similarly, $B_S$ maps a dialogue history, task schema $S$ and API client $C$ to a distribution over system turn annotations. In dialogue self-play (Figure \ref{fig:framework}c), a new turn annotation $a^i_j$ is iteratively sampled from each bot until either the user's goals are achieved and the user exits the dialogue with a ``bye()'' act, or a maximum number of turns are reached.

In our setup, $B_U$ is an agenda-based user simulator (\citet{schatzmann2007agenda}) with a modification that the action selection model is conditioned on the user profile in addition to the user goal and dialogue history. $B_S$ is modeled as a finite state machine (\citet{Hopcroft:2006:IAT:1177300}) which encodes a set of task-independent rules for constructing system turns, with each turn consisting of a \textit{response} frame which responds to the user's previous turn, and an \textit{initiate} frame which drives the dialogue forward through a predetermined sequence of sub-dialogues. For database querying applications, these sub-dialogues are: gather user preferences, query a database via an API, offer matching entities to the user, allow user to modify preferences or request more information about an entity, and finally complete the transaction (buying or reserving the entity). By exploring a range of parameter values for $B_U$ and $B_S$ and sampling a large number of outlines, dialogue self-play can generate a diverse set of dialogue outlines for the task.

Finally, a \textit{template utterance generator} maps each turn annotation to a template utterance, $a^i_j \rightarrow t^i_j$, using a domain-general grammar similar to the one described in \citet{wang2015building}. Alternatively, the developer can provide a list of templates to use for some or all of the dialogue frames, for example if they want more control over the language used in the system turns. The template utterances $t^i_j$ are an important bridge between the turn annotation $a^i_j$ and the corresponding natural language utterance $u^i_j$, since crowd workers may not understand the annotations if presented in symbolic form.

\subsection{Crowdsourced paraphrases}
To obtain a natural language dialogue from its outline, $F_2(o_i) \rightarrow d_i$, the framework employs crowd sourcing to paraphrase template utterances $t^i_j$ into more natural sounding utterances $u^i_j$. The paraphrase task is designed as a ``contextual rewrite'' task where a crowd worker sees the full dialogue template $t^i_1 \dots t^i_{n_i}$, and provides the natural language utterances $u^i_1 \dots u^i_{n_i}$ for each turn of the dialogue. A screenshot of the contextual rewrite task interface is provided in Figure \ref{fig:ewokpara}. This encourages the crowd worker to inject linguistic phenomena like coreference (``Reserve that restaurant'') and lexical entrainment into the utterances. We show the same outline to $K > 1$ crowdworkers to get more diversity of natural language utterances for the same dialogue, $\{u^i_j\}_k$.

Since $t^i_j$ and $u^i_j$ are paraphrases of each other, the annotations $a^i_j$ automatically apply to $u^i_j$, eliminating the need for an expensive annotation step. In practice, for a fraction of the utterances, the automatic annotation does not succeed either due to crowd workers not following instructions properly or if the utterance contains a paraphrase of a slot value, for example when the crowd worker rephrases ``between 5pm and 8pm'' as ``some time in the evening''. We employ a second round of crowdsourcing for validating the utterances. For each $u^i_j$, we ask two crowd workers if it has the same meaning as the corresponding template $t^i_j$, and we drop the utterance if either of the crowd workers say no. Dialogues which end up having no natural language utterance for at least one of the turns are dropped from the dataset. For the remaining utterances, slot values from the annotation $a^i_j$ are tagged in the utterance with substring match. If a slot value cannot be found automatically, we show it to two crowd workers and ask them to annotate the slot span. Alternatively, such annotation errors be detected and corrected by active learning (\citet{hakkani2002active,tur2003active}).

\subsection{Dataset expansion}
The rewrites $t \rightarrow \{u\}_k$ collected via the crowdsourcing step $F_2$ can be compiled into a map $L(a) \rightarrow \{u\}_k$. As an optional step, this map  could be leveraged to synthetically expand the dataset beyond what is economically feasible to collect via crowdsourcing. The self-play step $F_1$ can be executed to generate a larger set of outlines $O_S >> O$. For each turn annotation $a^i_j$ of $o_i \in O_S$, a natural language utterance is sampled\footnote{If $a^i_j \notin L$, then $o_i$ is dropped from $O_S$.} from $L(a^i_j)$ to create the corresponding dialogue $d_i \in D_S >> D$. Dialogues in the synthetic set $D_S$ could have utterances that were written by crowdworkers under a different context, so these dialogues are of a slightly lower quality.

\subsection{Model training}
The dialogues $d_i \in D$ (or $D_S$) have natural language turns along with annotations of dialogue acts, slot spans, dialogue state and API state for each turn. These labels are sufficient for training dialogue models from recent literature: either component-wise models for language understanding (\citet{bapna2017sequential}), state tracking (\citet{rastogi2017scalable}), dialogue policy (\citet{shah2016interactive}) and language generation (\citet{nayak2017plan}), or end-to-end models (\citet{wen2016network}). Further, we can construct a natural language user simulator by combining $U_B$ with $L(a)$, and use it to train end-to-end dialogue models with reinforcement learning (\citet{liu2017end}).

\section{User simulation and dialogue self-play}
Our framework hinges on having a generative model of a user that is reasonably close to actual users of the system. The motivation is that while it is hard to develop precise models of user behavior customized for every type of dialogue interaction, it is possible to create a domain-general user simulator that operates at a higher level of abstraction (dialogue acts) and encapsulates common patterns of user behavior for a broad class of dialogue tasks. Seeding the user simulator with a task-specific schema of intents, slot names and slot values allows the framework to generate a variety of dialogue flows tailored to that specific task. Developing a general user simulator targeting a broad class of tasks, for example database querying applications, has significant leverage as adding a new conversational pattern to the simulator benefits the outlines generated for dialogue interfaces to any database or third-party API.

Another concern with the use of a user simulator is that it restricts the generated dialogue flows to only those that are engineered into the user model. In comparison, asking crowd workers to converse without any restrictions could generate interesting dialogues that are not anticipated by the dialogue developer. Covering complex interactions is important when developing datasets to benchmark research aimed towards building human-level dialogue systems. However, we argue that for consumer-facing chatbots, the primary aim is reliable coverage of critical user interactions. Existing methods for developing chatbots with engineered finite state machines implicitly define a model of expected user behavior in the states and transitions of the system agent. A user simulator makes this user model explicit and is a more systematic approach for a dialogue developer to reason about the user behaviors handled by the agent. Similarly, having more control over the dialogue flows present in the dataset ensures that all and only expected user and system agent behaviors are present in the dataset. Our crowd sourcing setup obtains diverse natural language realizations of the abstract dialogue flows generated via self-play. A dialogue agent bootstrapped with such a dataset can be deployed in front of users with a guaranteed minimum task completion rate. Subsequently, the dialogue agent can be directly improved from real user interactions, for which crowdsourcing is anyways an imperfect substitute.

The self-play step also uses a system bot $B_S$ that generates valid system turns for a given task. Since our framework uses a rule-based bot which takes user dialogue acts as inputs and emits a neural network based dialogue agent that works with natural language utterances, the framework effectively distills expert knowledge into a learned neural network. The developer can customize the behavior of the neural agent by modifying the component rules of $B_S$. Further, the cost of developing $B_S$ can be amortized over a large class of dialogue tasks by building a domain-agnostic bot for handling a broad task like database querying applications, similar to $U_S$. Finally, in contrast to a rule-based bot, a neural dialogue agent is amenable to further improvement from direct user interactions via reinforcement learning (\citet{su2016line, liu2017end}), opening up the possibility of lifelong improvement in the quality of the dialogue agent.

\section{\label{sec:crowdsourcing}Crowdsourcing}
In the Wizard-of-Oz setting, the dialogue task specification is used to construct tasks by sampling slot values from the API client. A task is then shown to a pair of crowd workers who are asked to converse in natural language to complete the task. Subsequently, the collected dialogues are manually annotated with dialogue act and slot span labels for training dialogue models. This process is expensive as the two annotation tasks given to crowd workers in the WOz setting are difficult and therefore time consuming: identifying the dialogue acts of an utterance requires understanding the precise meaning of each dialogue act, and identifying all slot spans in an utterance requires checking the utterance against all slots in the schema. As a result, the crowdsourced annotations may need to be cleaned by an expert. In contrast, M2M significantly reduces the crowdsourcing expense by automatically annotating a majority of the dialogue turns and annotating the remaining turns with two simpler crowdsourcing tasks, ``Does this utterance contain this particular slot value?'' and ``Do these two utterances have the same meaning?'', which are more efficiently done by an average crowd worker.

Further, the lack of control over crowd workers' behavior in the Wizard-of-Oz setting can lead to dialogues that may not reflect the behavior of real users, for example if the crowd worker provides all constraints in a single turn. Such low-quality dialogues either need to be manually removed from the dataset or the crowd participants need to be given additional instructions or training to encourage better interactions (\citet{asri2017frames}). M2M avoids this issue by using dialogue self-play to systematically generate all usable dialogue outlines, and simplifying the crowdsourcing step to a dialogue paraphrase task.

\section{Datasets}
We are releasing\footnote{https://github.com/google-research-datasets/simulated-dialogue} two datasets totaling 3000 dialogues collected using M2M for the tasks of buying a movie ticket and reserving a restaurant table. (Table \ref{tab:datasets}). The datasets were collected by first generating outlines using dialogue self-play and then rewriting the template utterances using crowd sourcing.

\begin{table}[t]
\centering
\resizebox{0.98\linewidth}{!}{%
\begin{tabular}{|l|l|l|l|l|}
\hline
\multicolumn{1}{|c|}{\textbf{Dataset}} & \multicolumn{1}{c|}{\textbf{Slots}} & \multicolumn{1}{c|}{\textbf{Train}} & \multicolumn{1}{c|}{\textbf{Dev}} & \multicolumn{1}{c|}{\textbf{Test}} \\ \hline
Restaurant & \begin{tabular}[c]{@{}l@{}}price\_range, location,\\ restaurant\_name, category,\\ num\_people, date, time\end{tabular} & 1116 & 349 & 775 \\ \hline
Movie & \begin{tabular}[c]{@{}l@{}}theatre\_name, movie, date,\\ time, num\_people\end{tabular} & 384 & 120 & 264 \\ \hline
\end{tabular}%
}%
\caption{Dialogues collected with M2M.}
\label{tab:datasets}
\end{table}

\section{Evaluations}
We present some experiments with the M2M datasets to evaluate the M2M approach for collecting dialogue datasets and training conversational agents with that data.

\subsection{Dialogue diversity}
First we will investigate the claim that M2M leads to higher coverage of dialogue features in the dataset. We compare the M2M Restaurants training dialogues with the DSTC2 (\citet{henderson2013dialog}) training set which also deals with restaurant reservations (Table \ref{tab:metrics}). M2M compares favorably to DSTC2 on the ratio of unique unigrams and bigrams to total number of tokens in the dataset, which signifies a greater variety of surface forms as opposed to repeating the same words and phrases. Similarly, we count the number of unique ``transitions'' at the semantic frame level, defined as a pair of annotations $a_i, a_{i+1}$ of contiguous turns. This gives a measure of diversity of dialogue flows in the dataset. M2M has 3x the number of unique transitions per turn of the dataset. We also count unique ``subdialogues'', i.e. sequences of transitions $a_i, a_{i+1}, \dots, a_{i+k}$ for $k=\{3, 5\}$, and observe that M2M has fewer repetitions of subdialogues compared to DSTC2.

\begin{table}[t]
\centering
\resizebox{0.98\linewidth}{!}{%
\begin{tabular}{|l|c|c|}
\hline
\multicolumn{1}{|c|}{\textbf{Metric}} & \textbf{\begin{tabular}[c]{@{}c@{}}DSTC2\\ (Train)\end{tabular}} & \textbf{\begin{tabular}[c]{@{}c@{}}M2M Rest.\\ (Train)\end{tabular}} \\ \hline
Dialogues & 1611 & 1116 \\ \hline
Total turns & 11670 & 6188 \\ \hline
Total tokens & 199295 & 99932 \\ \hline
Avg. turns per dialogue & 14.49 & 11.09 \\ \hline
Avg. tokens per turn & 8.54 & 8.07 \\ \hline
\begin{tabular}[c]{@{}l@{}}Unique tokens /\\ Total tokens\end{tabular} & 0.0049 & 0.0092 \\ \hline
\begin{tabular}[c]{@{}l@{}}Unique bigrams /\\ Total tokens\end{tabular} & 0.0177 & 0.0670 \\ \hline
\begin{tabular}[c]{@{}l@{}}Unique transitions /\\ Total turns\end{tabular} & 0.0982 & 0.2646 \\ \hline
\begin{tabular}[c]{@{}l@{}}Unique subdialogues(k=3) / \\ Total subdialogues(n=3)\end{tabular} & 0.1831 & 0.3145 \\ \hline
\begin{tabular}[c]{@{}l@{}}Unique subdialogues(k=5) /\\ Total subdialogues(n=5)\end{tabular} & 0.5621 & 0.7061 \\ \hline
\begin{tabular}[c]{@{}l@{}}Unique full outlines /\\ Total dialogues\end{tabular} & 0.9243 & 0.9292 \\ \hline
\end{tabular}%
}%
\caption{Comparing DSTC2 and M2M Restaurants datasets on diversity of language and dialogue flows.}
\label{tab:metrics}
\end{table}

\subsection{Human evaluation of dataset quality}
For a subjective evaluation of the quality of the M2M datasets, we ran an experiment showing the final dialogues to crowd workers and asking them to rate each user and system turn between 1 to 5 on multiple dimensions. Figure \ref{fig:ewokquality} in the Appendix presents the interface shown to crowd workers for collecting the ratings. Each turn was shown to 3 crowd workers. Table \ref{tab:qualityeval} presents the mean and standard deviation of ratings aggregated over all turns of the datasets.

%\subsection{\label{sec:customization}Overnight experiment}
%We will now present a concrete example of collecting a dataset of dialogues in an entirely new task and training a dialogue agent to handle that task. We choose the task of booking a doctor's appointment on an online portal. Describe the steps: task schema with slots (doctor name, location, appointment date, reason). Show an example dialogue outline. Show the same dialogue after crowd paraphrases. Describe a task specific rule that could be added to the user simulator for this task.

\section{Related work and discussion}
We presented M2M, an extensible framework for rapidly bootstrapping goal-oriented conversational agents. Comparisons with the popular Dialog State Tracking Challenge 2 dataset (\citet{henderson2013dialog}) show that M2M can be leveraged for rapidly creating high-quality datasets for training conversational agents in arbitrary domains. A key benefit of our framework is that it is fully controllable via multiple knobs: the task schema, the scenario generator, the user profile and behavior, the system policy and the template generator. PyDial (\citet{ultes2017pydial}), an extensible open-source toolkit which provides domain-independent implementations of dialogue system modules, could be extended to support M2M by adding dialogue self-play functionality.

The user and system bots in this work are implemented using task-general rules so that any transactional or form-filling task could be handled with only the task schema. For more complex tasks, the developer can extend the user and system bots or the canonical utterance generator by adding their own rules. These components could also be replaced by machine learned generative models if available. Task Completion Platform (TCP) (\citet{crook2016task}) introduced a task configuration language for building goal-oriented dialogue interactions. The state update and policy modules of TCP could be used to implement bots that generate outlines for complex tasks.

\begin{table}[t]
\centering
\resizebox{0.68\linewidth}{!}{%
\begin{tabular}{lcc}
\hline
\multicolumn{1}{c}{\textbf{}} & \textbf{\begin{tabular}[c]{@{}c@{}}M2M\\ Restaurants\end{tabular}} & \textbf{\begin{tabular}[c]{@{}c@{}}M2M\\ Movies\end{tabular}} \\ \hline
\textbf{User:} & \multicolumn{1}{l}{} & \multicolumn{1}{l}{} \\
Natural & 4.66 (0.54) & 4.70 (0.49) \\ \hline
\textbf{System:} &  &  \\
Polite & 4.23 (0.62) & 4.27 (0.62) \\
Clear & 4.72 (0.52) & 4.75 (0.48) \\
Optimal & 4.26 (0.76) & 4.32 (0.75) \\ \hline
\end{tabular}%
}%
\caption{Human evaluation of dialogues collected with M2M. Average of crowd worker scores (from 1 to 5) for user and system turns (standard deviation in brackets).}
\label{tab:qualityeval}
\end{table}

ParlAI (\citet{miller2017parlai}), a dialogue research software platform, provides easy integration with crowd sourcing for data collection and evaluation. However, the crowd sourcing tasks are open-ended and may result in lower quality dialogues as described in Section \ref{sec:crowdsourcing}. The crowd sourcing tasks in M2M are configured to be at a suitable difficulty level for crowd workers as they are neither open-ended nor too restrictive. The crowd workers are asked to paraphrase utterances instead of coming up with completely new ones. 

\section*{Acknowledgements}
We thank Georgi Nikolov, Amir Fayazi, Anna Khasin and Grady Simon for valuable support in design, implementation and evaluation of M2M.

%\bibliography{naaclhlt2018}
\bibliography{naaclhlt2018}

\begin{thebibliography}{}
\expandafter\ifx\csname natexlab\endcsname\relax\def\natexlab#1{#1}\fi

\bibitem[{Asri et~al.(2017)Asri, Schulz, Sharma, Zumer, Harris, Fine, Mehrotra,
  and Suleman}]{asri2017frames}
Layla~El Asri, Hannes Schulz, Shikhar Sharma, Jeremie Zumer, Justin Harris,
  Emery Fine, Rahul Mehrotra, and Kaheer Suleman. 2017.
\newblock Frames: A corpus for adding memory to goal-oriented dialogue systems.
\newblock {\em arXiv preprint arXiv:1704.00057\/} .

\bibitem[{Bapna et~al.(2017)Bapna, Tur, Hakkani-Tur, and
  Heck}]{bapna2017sequential}
Ankur Bapna, Gokhan Tur, Dilek Hakkani-Tur, and Larry Heck. 2017.
\newblock Sequential dialogue context modeling for spoken language
  understanding.
\newblock In {\em Proc. of SIGDIAL\/}.

\bibitem[{Crook et~al.(2016)Crook, Marin, Agarwal, Aggarwal, Anastasakos,
  Bikkula, Boies, Celikyilmaz, Chandramohan, Feizollahi et~al.}]{crook2016task}
PA~Crook, A~Marin, V~Agarwal, K~Aggarwal, T~Anastasakos, R~Bikkula, D~Boies,
  A~Celikyilmaz, S~Chandramohan, Z~Feizollahi, et~al. 2016.
\newblock Task completion platform: A self-serve multi-domain goal oriented
  dialogue platform.
\newblock {\em NAACL HLT 2016\/} page~47.

\bibitem[{Hakkani-T{\"u}r et~al.(2002)Hakkani-T{\"u}r, Riccardi, and
  Gorin}]{hakkani2002active}
Dilek Hakkani-T{\"u}r, Giuseppe Riccardi, and Allen Gorin. 2002.
\newblock Active learning for automatic speech recognition.
\newblock In {\em Acoustics, Speech, and Signal Processing (ICASSP), 2002 IEEE
  International Conference on\/}. IEEE, volume~4, pages IV--3904.

\bibitem[{Henderson et~al.(2013)Henderson, Thomson, and
  Williams}]{henderson2013dialog}
Matthew Henderson, Blaise Thomson, and Jason Williams. 2013.
\newblock Dialog state tracking challenge 2 \& 3.
\newblock \url{http://camdial.org/~mh521/dstc/}.

\bibitem[{Hopcroft et~al.(2006)Hopcroft, Motwani, and
  Ullman}]{Hopcroft:2006:IAT:1177300}
John~E. Hopcroft, Rajeev Motwani, and Jeffrey~D. Ullman. 2006.
\newblock {\em Introduction to Automata Theory, Languages, and Computation (3rd
  Edition)\/}.
\newblock Addison-Wesley Longman Publishing Co., Inc., Boston, MA, USA.

\bibitem[{Liu et~al.(2017)Liu, Tur, Hakkani-Tur, Shah, and Heck}]{liu2017end}
Bing Liu, Gokhan Tur, Dilek Hakkani-Tur, Pararth Shah, and Larry Heck. 2017.
\newblock End-to-end optimization of task-oriented dialogue model with deep
  reinforcement learning.
\newblock In {\em NIPS Conversational AI Workshop\/}.

\bibitem[{Miller et~al.(2017)Miller, Feng, Fisch, Lu, Batra, Bordes, Parikh,
  and Weston}]{miller2017parlai}
Alexander~H Miller, Will Feng, Adam Fisch, Jiasen Lu, Dhruv Batra, Antoine
  Bordes, Devi Parikh, and Jason Weston. 2017.
\newblock Parlai: A dialog research software platform.
\newblock {\em arXiv preprint arXiv:1705.06476\/} .

\bibitem[{Nayak et~al.(2017)Nayak, Hakkani-Tur, Walker, and
  Heck}]{nayak2017plan}
Neha Nayak, Dilek Hakkani-Tur, Marilyn Walker, and Larry Heck. 2017.
\newblock To plan or not to plan? discourse planning in slot-value informed
  sequence to sequence models for language generation.
\newblock In {\em Proc. of Interspeech\/}.

\bibitem[{Rastogi et~al.(2017)Rastogi, Hakkani-Tur, and
  Heck}]{rastogi2017scalable}
Abhinav Rastogi, Dilek Hakkani-Tur, and Larry Heck. 2017.
\newblock Scalable multi-domain dialogue state tracking.
\newblock In {\em Proc. of IEEE ASRU\/}.

\bibitem[{Schatzmann et~al.(2007)Schatzmann, Thomson, Weilhammer, Ye, and
  Young}]{schatzmann2007agenda}
Jost Schatzmann, Blaise Thomson, Karl Weilhammer, Hui Ye, and Steve Young.
  2007.
\newblock Agenda-based user simulation for bootstrapping a pomdp dialogue
  system.
\newblock In {\em Human Language Technologies 2007: The Conference of the North
  American Chapter of the Association for Computational Linguistics; Companion
  Volume, Short Papers\/}. Association for Computational Linguistics, pages
  149--152.

\bibitem[{Shah et~al.(2016)Shah, Hakkani-T{\"u}r, and
  Heck}]{shah2016interactive}
Pararth Shah, Dilek Hakkani-T{\"u}r, and Larry Heck. 2016.
\newblock Interactive reinforcement learning for task-oriented dialogue
  management.
\newblock In {\em NIPS Deep Learning for Action and Interaction Workshop\/}.

\bibitem[{Silver et~al.(2016)Silver, Huang, Maddison, Guez, Sifre, Van
  Den~Driessche, Schrittwieser, Antonoglou, Panneershelvam, Lanctot
  et~al.}]{silver2016mastering}
David Silver, Aja Huang, Chris~J Maddison, Arthur Guez, Laurent Sifre, George
  Van Den~Driessche, Julian Schrittwieser, Ioannis Antonoglou, Veda
  Panneershelvam, Marc Lanctot, et~al. 2016.
\newblock Mastering the game of go with deep neural networks and tree search.
\newblock {\em Nature\/} 529(7587):484--489.

\bibitem[{Silver et~al.(2017)Silver, Hubert, Schrittwieser, Antonoglou, Lai,
  Guez, Lanctot, Sifre, Kumaran, Graepel et~al.}]{silver2017mastering}
David Silver, Thomas Hubert, Julian Schrittwieser, Ioannis Antonoglou, Matthew
  Lai, Arthur Guez, Marc Lanctot, Laurent Sifre, Dharshan Kumaran, Thore
  Graepel, et~al. 2017.
\newblock Mastering chess and shogi by self-play with a general reinforcement
  learning algorithm.
\newblock {\em arXiv preprint arXiv:1712.01815\/} .

\bibitem[{Su et~al.(2016)Su, Ga{\v{s}}i{\'c}, Mrk{\v{s}}i{\'c}, Rojas-Barahona,
  Ultes, Vandyke, Wen, and Young}]{su2016line}
PH~Su, M~Ga{\v{s}}i{\'c}, N~Mrk{\v{s}}i{\'c}, L~Rojas-Barahona, S~Ultes,
  D~Vandyke, TH~Wen, and S~Young. 2016.
\newblock On-line active reward learning for policy optimisation in spoken
  dialogue systems.
\newblock In {\em 54th Annual Meeting of the Association for Computational
  Linguistics, ACL 2016-Long Papers\/}. volume~4, pages 2431--2441.

\bibitem[{Tur et~al.(2003)Tur, Schapire, and Hakkani-Tur}]{tur2003active}
Gokhan Tur, Robert~E Schapire, and Dilek Hakkani-Tur. 2003.
\newblock Active learning for spoken language understanding.
\newblock In {\em Acoustics, Speech, and Signal Processing, 2003.
  Proceedings.(ICASSP'03). 2003 IEEE International Conference on\/}. IEEE,
  volume~1, pages I--I.

\bibitem[{Ultes et~al.(2017)Ultes, Barahona, Su, Vandyke, Kim, Casanueva,
  Budzianowski, Mrk{\v{s}}i{\'c}, Wen, Gasic et~al.}]{ultes2017pydial}
Stefan Ultes, Lina M~Rojas Barahona, Pei-Hao Su, David Vandyke, Dongho Kim,
  Inigo Casanueva, Pawe{\l} Budzianowski, Nikola Mrk{\v{s}}i{\'c}, Tsung-Hsien
  Wen, Milica Gasic, et~al. 2017.
\newblock Pydial: A multi-domain statistical dialogue system toolkit.
\newblock {\em Proceedings of ACL 2017, System Demonstrations\/} pages 73--78.

\bibitem[{Wang et~al.(2015)Wang, Berant, Liang et~al.}]{wang2015building}
Yushi Wang, Jonathan Berant, Percy Liang, et~al. 2015.
\newblock Building a semantic parser overnight.
\newblock {\em ACL\/} .

\bibitem[{Wen et~al.(2016)Wen, Vandyke, Mrksic, Gasic, Rojas-Barahona, Su,
  Ultes, and Young}]{wen2016network}
Tsung-Hsien Wen, David Vandyke, Nikola Mrksic, Milica Gasic, Lina~M
  Rojas-Barahona, Pei-Hao Su, Stefan Ultes, and Steve Young. 2016.
\newblock A network-based end-to-end trainable task-oriented dialogue system.
\newblock {\em ACL\/} .

\bibitem[{Zhong et~al.(2017)Zhong, Xiong, and Socher}]{zhong2017seq2sql}
Victor Zhong, Caiming Xiong, and Richard Socher. 2017.
\newblock Seq2sql: Generating structured queries from natural language using
  reinforcement learning.
\newblock {\em arXiv preprint arXiv:1709.00103\/} .

\end{thebibliography}
\bibliographystyle{acl_natbib}

\appendix

\section{Supplemental Material}
\label{sec:supplemental}
Table \ref{tab:dacts} lists the dialogue acts used in our setup. The dialogue acts are based on the Cambridge dialogue act set. Table \ref{tab:sampledialogue} presents a full dialogue outline and corresponding paraphrase for a dialogue spanning two interdependent tasks, where the user wants to first buy movie tickets and then reserve a restaurant table for dinner after the movie.

\begin{figure*}
  \caption{Contextual rewrite task interface for paraphrasing a dialogue outline with natural language.}
  \includegraphics[width=\textwidth]{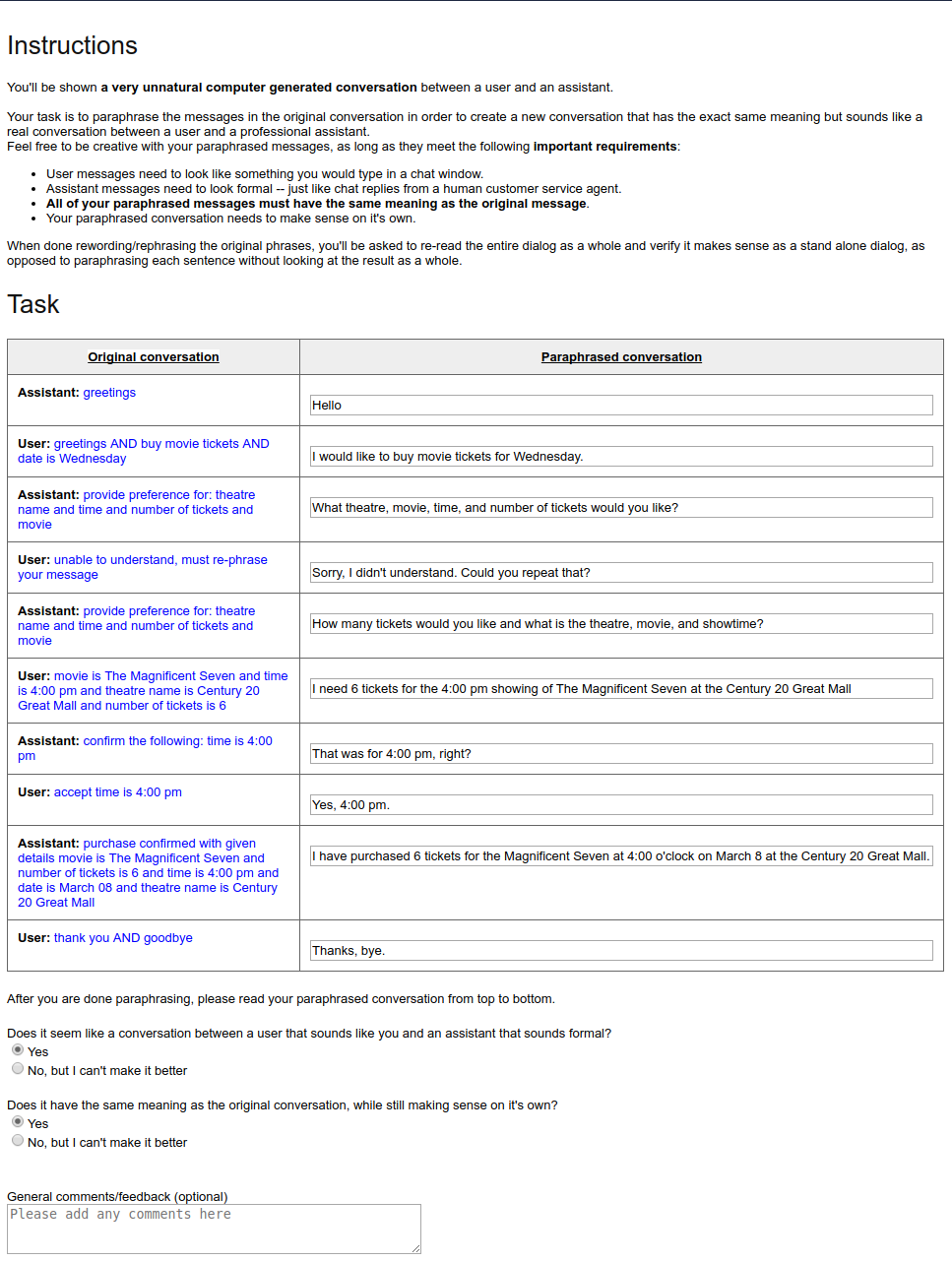}
  \label{fig:ewokpara}
\end{figure*}

\begin{figure*}
  \caption{Dialogue quality evaluation task interface for rating the user and system turns of completed dialogues.}
  \includegraphics[width=\textwidth]{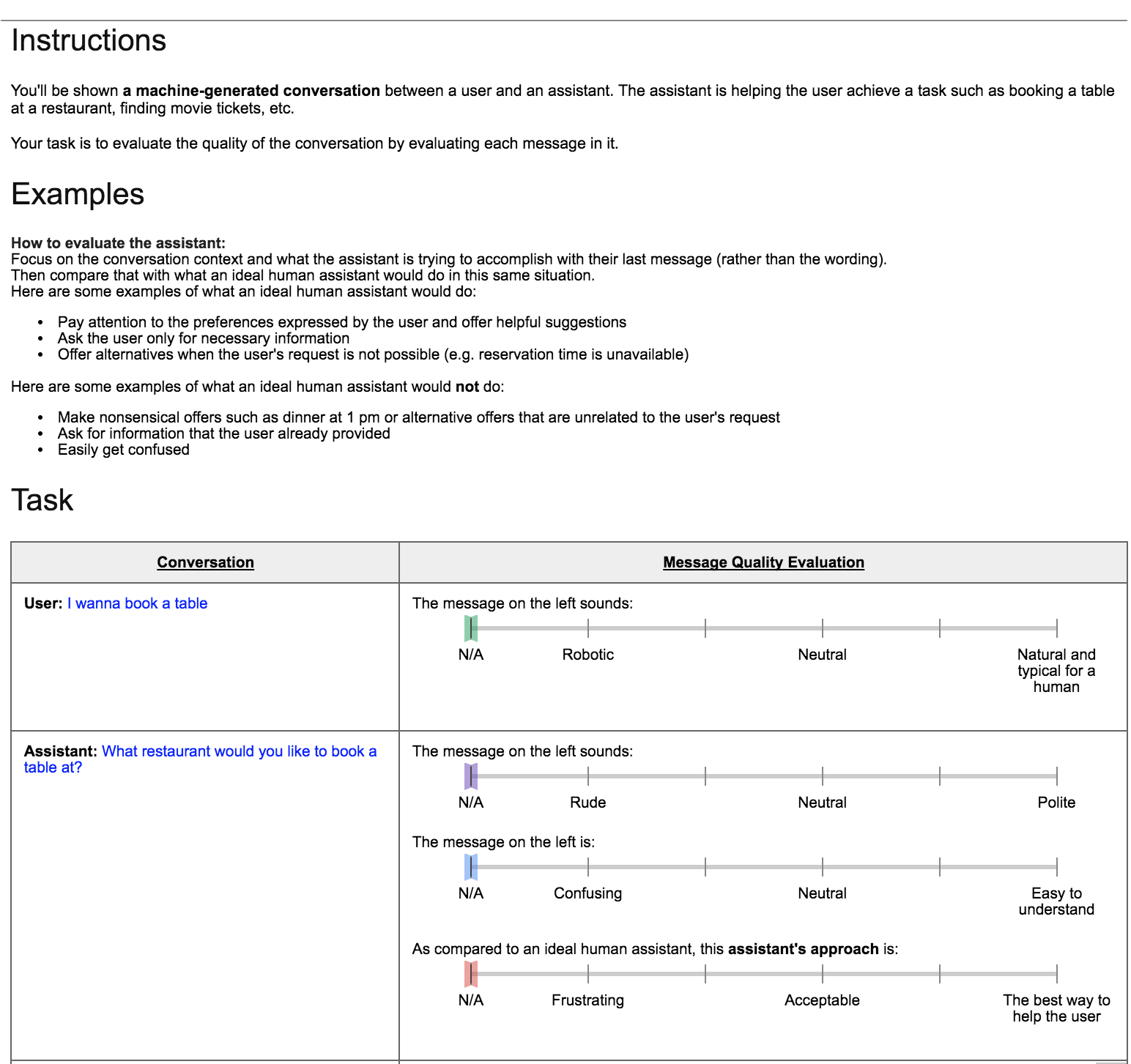}
  \label{fig:ewokquality}
\end{figure*}

Figure \ref{fig:ewokpara} presents the interface shown to crowd workers for the dialogue rewrite task, and includes a sample dialogue outline (consisting of template utterances) and its paraphrase into natural language. Figure \ref{fig:ewokquality} presents the interface shown to crowd workers for evaluating the quality of dialogues collected with M2M.

\begin{table*}[t]
\centering
\caption{List of dialogue acts.}
\label{tab:dacts}
\resizebox{0.98\linewidth}{!}{%
\begin{tabular}{lcl}
\hline
\multicolumn{1}{c}{\textbf{Dialogue Act}} & \textbf{Speaker} & \multicolumn{1}{c}{\textbf{Description}} \\ \hline
GREETING & User/System & Greet the other speaker \\
INFORM & User/System & Inform a slot value \\
CONFIRM & User/System & Ask the other speaker to confirm a given slot value \\
REQUEST & User/System & Ask for the value of a slot \\
REQUEST\_ALTS & User & Ask for more alternatives \\
OFFER & System & Offer a database entity to the user \\
SELECT & System & Offer more than one database entity to the user \\
AFFIRM & User/System & Agree to something said by the other speaker \\
NEGATE & User/System & Disagree to something said by the other speaker \\
NOTIFY\_SUCCESS & System & Notify the user of a successful event, e.g. a booking is complete \\
NOTIFY\_FAILURE & System & Notify the user of a failure event, e.g. a booking isn't available \\
THANK\_YOU & User/System & Thank the other speaker \\
GOOD\_BYE & User/System & Say goodbye to the other speaker \\
CANT\_UNDERSTAND & User/System & Tell the other speaker that their utterance was not understood \\
OTHER & User & Unknown utterance type \\ \hline
\end{tabular}%
}%
\end{table*}

\begin{table*}[ht]
\centering
\caption{Sample multi-domain dialogue outline and paraphrase.}
\label{tab:sampledialogue}
\resizebox{0.98\linewidth}{!}{%
\begin{tabular}{|l|l|l|}
\hline
\multicolumn{2}{|c|}{\textbf{Outline}} & \multicolumn{1}{c|}{\textbf{Paraphrase}} \\ \hline
\multicolumn{1}{|c|}{\textbf{Annotation ($a_i$)}} & \multicolumn{1}{c|}{\textbf{Template utterances ($t_i$)}} & \multicolumn{1}{c|}{\textbf{NL utterances ($u_i$)}} \\ \hline
S: greeting() & Greeting. & Hi, how can I help you? \\ \hline
\begin{tabular}[c]{@{}l@{}}U: inform(intent=book\_movie, \\ name=Inside Out, date=tomorrow,\\ num\_tickets=2)\end{tabular} & \begin{tabular}[c]{@{}l@{}}Book movie with name is \\ Inside Out and date is tomorrow\\ and num tickets is 2.\end{tabular} & \begin{tabular}[c]{@{}l@{}}I want to buy 2 tickets for Inside\\ Out for tomorrow.\end{tabular} \\ \hline
S: ack() request(time) & OK. Provide time. & \begin{tabular}[c]{@{}l@{}}Alright. What time would you like\\ to see the movie?\end{tabular} \\ \hline
U: inform(time=evening) & Time is evening. & \begin{tabular}[c]{@{}l@{}}Anytime during the evening works\\ for me.\end{tabular} \\ \hline
\begin{tabular}[c]{@{}l@{}}S: offer(theatre=Cinemark 16,\\ time=6pm)\end{tabular} & \begin{tabular}[c]{@{}l@{}}Offer theatre is Cinemark 16 and\\ time is 6pm.\end{tabular} & \begin{tabular}[c]{@{}l@{}}How about the 6pm show at\\ Cinemark 16?\end{tabular} \\ \hline
U: affirm() & Agree. & That sounds good. \\ \hline
S: notify\_success() & Reservation confirmed. & Your tickets have been booked! \\ \hline
\begin{tabular}[c]{@{}l@{}}U: inform(intent=find\_restaurant,\\ meal=dinner, location=near the\\ theatre)\end{tabular} & \begin{tabular}[c]{@{}l@{}}Find restaurant with meal is dinner\\ and location is near the theatre.\end{tabular} & \begin{tabular}[c]{@{}l@{}}I want to get dinner at a restaurant\\ near the theatre.\end{tabular} \\ \hline
S: request(cuisine, price\_range) & Provide cuisine and price range. & \begin{tabular}[c]{@{}l@{}}Do you have any preference for\\ cuisine or price range?\end{tabular} \\ \hline
\begin{tabular}[c]{@{}l@{}}U: inform(cuisine=DontCare,\\ price\_range=moderate, rating=high)\end{tabular} & \begin{tabular}[c]{@{}l@{}}Cuisine is I don't care and price\\ range is moderate and rating is high.\end{tabular} & \begin{tabular}[c]{@{}l@{}}I'm fine with any cuisine. Look for\\ something moderately priced, but\\ make sure it has a high rating.\end{tabular} \\ \hline
\begin{tabular}[c]{@{}l@{}}S: select(restaurant=\{First Wok,\\ Lucy's Grill\}, location=near the\\ theatre)\end{tabular} & \begin{tabular}[c]{@{}l@{}}Select restaurant from First Wok,\\ Lucy's Grill with location is near the\\ theatre.\end{tabular} & \begin{tabular}[c]{@{}l@{}}First Wok and Lucy's Grill are\\ some good options near the\\ theatre.\end{tabular} \\ \hline
\begin{tabular}[c]{@{}l@{}}U: inform(intent=reserve\_restaurant,\\ restaurant=First Wok, time=after the\\ movie)\end{tabular} & \begin{tabular}[c]{@{}l@{}}Reserve restaurant with restaurant is\\ First Wok and time is after the\\ movie.\end{tabular} & \begin{tabular}[c]{@{}l@{}}First Wok sounds perfect. Can you\\ reserve a table there for dinner \\ after the movie?\end{tabular} \\ \hline
\begin{tabular}[c]{@{}l@{}}S: ack() confirm(restaurant=First\\ Wok, time=8pm, num\_people=2)\end{tabular} & \begin{tabular}[c]{@{}l@{}}OK. Confirm restaurant is First Wok\\ and time is 8pm and num people is 2.\end{tabular} & \begin{tabular}[c]{@{}l@{}}Sure. Please confirm that it is a table\\ for 2 at First Wok for 8pm.\end{tabular} \\ \hline
U: affirm() & Agree. & That is correct. \\ \hline
S: notify\_success() & Reservation confirmed. & Your table has been reserved. \\ \hline
U: thank\_you() good\_bye() & Thank you and good bye. & Thanks! That's all for now. \\ \hline
\end{tabular}%
}%
\end{table*}

\newpage

\end{document}